\documentclass[twoside,11pt]{article}

\usepackage{blindtext}
\usepackage{amsmath,amssymb,mathtools,bm}
\usepackage{booktabs}
\usepackage{microtype}
\usepackage{graphicx}
\usepackage{url}

\usepackage[abbrvbib, preprint]{jmlr2e}

\usepackage{jmlr2e}

\usepackage{lastpage}
\jmlrheading{XX}{2026}{1-\pageref{LastPage}}{XX/XX}{XX/XX}{XX-XXXX}{Soumyadip Sarkar}

\ShortHeadings{MiniTensor: Lightweight, High-Performance Tensor Ops}{Soumyadip Sarkar}
\firstpageno{1}

\begin{document}

\title{MiniTensor: A Lightweight, High-Performance Tensor Operations Library}

\author{\name Soumyadip Sarkar \email contact.soumyadipsarkar@gmail.com \\
       \addr Independent Researcher}

\editor{My editor}

\maketitle

\begin{abstract}
We present \emph{MiniTensor}, an open source tensor operations library that focuses on minimalism, correctness, and performance. MiniTensor exposes a familiar PyTorch-like Python API while it executes performance critical code in a Rust engine. The core supports dense $n$\,dimensional tensors, broadcasting, reductions, matrix multiplication, reverse mode automatic differentiation, a compact set of neural network layers, and standard optimizers. In this paper, we describe the design of MiniTensor's architecture, including its efficient memory management, dynamic computation graph for gradients, and integration with Python via PyO3. We also compare the install footprint with PyTorch and TensorFlow to demonstrate that MiniTensor achieves a package size of only a few megabytes, several orders of magnitude smaller than mainstream frameworks, while preserving the essentials needed for research and development on CPUs. The repository can be found at \url{https://github.com/neuralsorcerer/minitensor}
\end{abstract}

\begin{keywords}
Tensor calculus, reverse mode automatic differentiation, deep learning, Rust, PyO3
\end{keywords}

\section{Introduction}
Modern deep learning and numerical computing rely heavily on tensor operation libraries. Popular frameworks like \textit{PyTorch} \citep{Paszke2019} and \textit{TensorFlow} \citep{Abadi2016} provide optimized tensor computations and automatic differentiation (autodiff), enabling rapid development of neural networks. However, these frameworks are extremely complex and heavyweight: they consist of millions of lines of code and large binary footprints, which can pose challenges in terms of understanding, maintainability, and deployment size. In many scenarios, a lightweight library that offers essential functionality with lower overhead is desirable. For example, researchers often benefit from minimal frameworks for education, quick prototyping, or embedding into applications where full-scale frameworks are impractical.

In this paper, we present \textbf{MiniTensor}, a lightweight yet high-performance tensor operations library. MiniTensor is designed to offer the core features of a deep learning framework such as multi-dimensional arrays (tensors), basic arithmetic and linear algebra operations, automatic differentiation for computing gradients, neural network building blocks, and optimization algorithms all within a minimal codebase. The key characteristics of MiniTensor include:
\begin{itemize}
    \item \textbf{PyTorch-like API:} MiniTensor adopts an imperative, eager execution style API similar to PyTorch, which lowers the learning curve for users familiar with existing frameworks. Tensors, neural network modules, and optimizers in MiniTensor can be used with a syntax and semantics inspired by PyTorch, making it easy to pick up.
    \item \textbf{Rust Backend for Performance:} The computational engine of MiniTensor is implemented in Rust, a systems programming language known for memory safety and high performance. Although in terms of optimizations and performance as of now both \textit{PyTorch} \citep{Paszke2019} and \textit{TensorFlow} \citep{Abadi2016} beats MiniTensor.
    \item \textbf{Lightweight Design:} This is the main core feature. MiniTensor focuses on essential features, avoiding unnecessary bloating. The built binary package is on the order of a few megabytes. In comparison, mainstream frameworks often have binary distributions hundreds of megabytes in size. This lightweight nature makes MiniTensor easy to integrate, audit, and extend.
    \item \textbf{Automatic Differentiation:} MiniTensor includes a built-in reverse-mode automatic differentiation engine \citep{Baydin2018}, enabling gradient computation for use in optimization and training of models. The autodiff system is transparent to the user, any sequence of tensor operations can be followed by a \texttt{backward()} call to compute gradients, similar to Autograd in PyTorch.
    \item \textbf{Neural Network Modules and Optimizers:} Despite its small size, MiniTensor provides a suite of common neural network components. These include layer abstractions (fully-connected layers, convolutional layers, activation functions, etc.) and loss functions, as well as optimizers like Stochastic Gradient Descent (SGD) and Adam \citep{Kingma2015}. This allows users to construct and train neural networks end-to-end using MiniTensor.
    \item \textbf{Python Integration and NumPy Interoperability:} MiniTensor is distributed with Python bindings via PyO3, making it usable as a regular Python library (installable via \texttt{pip}). It offers seamless conversion to and from NumPy \citep{Harris2020} arrays without copying data, allowing users to leverage existing scientific Python ecosystems.
\end{itemize}

We also verify that the published wheel is small compared to PyTorch and TensorFlow wheels, which frequently exceed hundreds of megabytes on common platforms \citep{PyPI-Torch,PyPI-TensorFlow,PyPI-MiniTensor}.

\paragraph{Notation.}
Bold lower case denotes vectors, bold upper case denotes matrices, and calligraphic symbols denote graphs. Given a scalar loss $L$, we write $\nabla_{\bm{\theta}}L$ for parameter gradients, $\circ$ for function composition, and $\odot$ for elementwise product.

\section{Related Work}
The idea of providing efficient tensor computations with autodiff is at the core of many modern libraries. \textbf{PyTorch} \citep{Paszke2019} is one of the most widely used frameworks; it introduced an imperative execution model with dynamic computation graphs, which greatly improved flexibility in building neural networks. PyTorch is implemented primarily in C++ (with Python bindings) and optimized for performance on both CPU and GPU. \textbf{TensorFlow} \citep{Abadi2016}, initially released by Google, took a different approach with static computation graphs and a declarative style (although later versions introduced eager execution mode). These frameworks provide industrial-strength performance and a vast array of features (from distributed training to visualization), but as a consequence, they are large and complex, often difficult for a single developer to fully comprehend or modify. For instance, the core of PyTorch’s autograd spans dozens of C++ source files and uses a custom intermediate representation for graph nodes \citep{Paszke2019}. 

There has been interest in simpler systems for automatic differentiation and deep learning. \textbf{Autograd} \citep{Baydin2018} (not to be confused with PyTorch's component of the same name) was an early Python library that could automatically compute gradients for NumPy operations using function overloading. \textbf{JAX} is a more recent system by Google that uses a tracing JIT compiler and XLA to optimize NumPy-based computations and provide autodiff. JAX can achieve very high performance by just-in-time compiling computational graphs, but it departs from the pure eager execution model and can be less intuitive for beginners.

In the quest for minimalism, some projects have demonstrated that a deep learning framework can be implemented in only a few hundred lines of code. \textbf{micrograd} \citep{karpathy2020micrograd} is an educational pure-Python autograd engine with a tiny codebase. \textbf{tinygrad} \citep{hotz2020tinygrad} is another minimalist deep learning library; it provides core tensor operations and training of simple models with a very small code size (and even has experimental GPU support). However, these ultra-light frameworks written in Python trade performance for simplicity: due to the overhead of Python loops and the GIL (Global Interpreter Lock), their execution can be orders of magnitude slower than optimized libraries in C++ or Rust.

MiniTensor seeks to occupy an interesting middle ground in this landscape. It is inspired by the dynamic graph approach of PyTorch and inherits a similar user interface, but it strives to remain as lightweight as projects like tinygrad in terms of code simplicity and size. By implementing the core in Rust, MiniTensor avoids the performance penalty typically associated with pure Python minimal libraries, and can approach the speed of production-grade frameworks on CPU tasks.

\section{Architecture}

MiniTensor adopts a three layer design: a Python API, a PyO3 bindings layer, and a Rust execution engine \citep{MiniTensorRepo}.

\subsection{Tensors and Primitive Operations}

A tensor is an $n$\,dimensional array with shape $\bm{s}=(s_1,\dots,s_n)$ and contiguous row major layout. The engine stores a typed buffer and lightweight metadata (shape and optional strides). For $x\in\mathbb{R}^{m\times k}$ and $W\in\mathbb{R}^{d\times k}$, matrix multiplication computes
\begin{equation}
\label{eq:matmul}
Y = X W^\top,\qquad Y\in\mathbb{R}^{m\times d}.
\end{equation}
Elementwise operations map as $z_i = f(x_i,y_i)$ for a binary $f$. Reductions implement linear functionals such as $\mathrm{sum}(x)=\sum_i x_i$ and averages such as $\mathrm{mean}(x)=\tfrac1N\sum_i x_i$.

Broadcasting follows NumPy and PyTorch rules \citep{Harris2020,Paszke2019}. For shapes that match after left padding singleton dimensions, the engine virtually expands along dimensions with size one. Consider $x\in\mathbb{R}^{b\times d}$ and $b\in\mathbb{R}^{d}$, then broadcasting computes $(x+b)_{ij} = x_{ij}+b_j$ without materializing $b$ across the batch dimension.

\subsection{Reverse Mode Automatic Differentiation}

MiniTensor records a computation graph $\mathcal{G}$ during the forward pass whenever a tensor requires gradients. Each node stores references to its parents and a \emph{local pullback} that maps an output cotangent to input cotangents. Let $L:\mathbb{R}^{n}\to\mathbb{R}$ be a scalar loss and let $y = f(x)$ define a differentiable primitive. Reverse mode propagates a seed $\bar{y}=\partial L/\partial y$ through the vector Jacobian product
\begin{equation}
\label{eq:vjp}
\bar{x} \;=\; \bar{y}\, J_f(x),\qquad J_f(x)=\frac{\partial f(x)}{\partial x},
\end{equation}
which is the transpose of the forward mode Jacobian vector product. For compositions $y=f_k\circ\cdots\circ f_1(x)$, the chain rule yields
\begin{equation}
\label{eq:chain}
\bar{x} \;=\; \bar{y}\, J_{f_k}(x_{k-1}) \cdots J_{f_{1}}(x_0).
\end{equation}
Reverse mode computes all parameter gradients $\nabla_{\bm{\theta}} L$ with time complexity proportional to a small constant multiple of the forward cost for scalar $L$ \citep{Baydin2018}.

\paragraph{Examples of local pullbacks.}
For $z=x+y$, the pullbacks satisfy $\bar{x}\mathrel{+{=}}\bar{z}$ and $\bar{y}\mathrel{+{=}}\bar{z}$. For Hadamard product $z=x\odot y$, they satisfy $\bar{x}\mathrel{+{=}}\bar{z}\odot y$ and $\bar{y}\mathrel{+{=}}\bar{z}\odot x$. For matrix multiplication in \eqref{eq:matmul}, the pullbacks are
\begin{equation}
\bar{X}\mathrel{+{=}}\bar{Y}\, W,\qquad \bar{W}\mathrel{+{=}}\bar{Y}^\top X.
\end{equation}

\subsection{Neural Network Layers, Losses, and Optimizers}

MiniTensor implements a small set of layers that cover common research and educational workloads \citep{MiniTensorRepo}.

\paragraph{Dense layer.}
Let $x\in\mathbb{R}^{b\times d_{\text{in}}}$, weight $W\in\mathbb{R}^{d_{\text{out}}\times d_{\text{in}}}$, and bias $b\in\mathbb{R}^{d_{\text{out}}}$. The forward map reads
\begin{equation}
\label{eq:dense}
\mathrm{Dense}(x;W,b) = x W^\top + \mathbf{1}\, b^\top,
\end{equation}
with gradients as above.

\paragraph{Convolution.}
For a 2D convolution with stride $s$ and zero padding $p$, the output at channel $c$ and spatial index $(i,j)$ is
\begin{equation}
y_{c,i,j} \;=\; \sum_{c'} \sum_{u=1}^{K_h}\sum_{v=1}^{K_w} w_{c,c',u,v}\; x_{c',\,i s + u - p,\, j s + v - p}.
\end{equation}
The engine implements the standard pullbacks with respect to $x$ and $w$.

\paragraph{Nonlinearities.}
MiniTensor provides ReLU, Sigmoid, Tanh, and GELU with the usual derivatives, for example $\partial\mathrm{ReLU}(x)/\partial x = \mathbb{I}\{x>0\}$.

\paragraph{Normalization and regularization.}
Batch normalization on activations $x\in\mathbb{R}^{b\times d}$ computes
\begin{equation}
\mu = \tfrac1b \sum_{i=1}^b x_i,\qquad \sigma^2 = \tfrac1b \sum_{i=1}^b (x_i-\mu)^2,\qquad
\mathrm{BN}_\gamma^\beta(x) = \gamma \odot \frac{x-\mu}{\sqrt{\sigma^2+\varepsilon}} + \beta,
\end{equation}
with learnable scale $\gamma$ and shift $\beta$ \citep{Ioffe2015}. Dropout applies an elementwise Bernoulli mask during training.

\paragraph{Losses.}
For multiclass classification with logits $z\in\mathbb{R}^{b\times C}$ and labels $y\in\{1,\dots,C\}^b$, cross entropy reads
\begin{equation}
\mathcal{L}_{\text{CE}}(z,y) = -\frac1b \sum_{i=1}^b \log \frac{\exp(z_{i,y_i})}{\sum_{c=1}^C \exp(z_{i,c})}.
\end{equation}
Gradients follow from the softmax derivative. Mean squared error implements $\mathcal{L}_{\text{MSE}}(x,\hat{x})=\tfrac1N\sum_i (x_i-\hat{x}_i)^2$.

\paragraph{Optimizers.}
Stochastic gradient descent with momentum maintains velocity $v_t$ and updates
\begin{equation}
v_t = \mu v_{t-1} + \nabla_{\bm{\theta}} L_t + \lambda \bm{\theta}_t,\qquad
\bm{\theta}_{t+1}=\bm{\theta}_t - \eta v_t.
\end{equation}
Adam maintains first and second moment estimates $m_t$ and $v_t$ with debiasing \citep{Kingma2015}:
\begin{equation}
m_t = \beta_1 m_{t-1} + (1-\beta_1) g_t,\quad
v_t = \beta_2 v_{t-1} + (1-\beta_2) g_t^2,\quad
\bm{\theta}_{t+1} = \bm{\theta}_t - \eta \frac{\hat{m}_t}{\sqrt{\hat{v}_t}+\epsilon}.
\end{equation}
RMSprop uses an exponential average of squared gradients and scales steps by $(v_t+\epsilon)^{-1/2}$ \citep{Tieleman2012}.

\subsection{Bindings and Interoperability}

MiniTensor exposes a Python module through PyO3. The bindings convert between Python objects and Rust buffers with zero copy where possible, for example when a tensor views a compatible NumPy array or returns a view to Python \citep{Pyo3Guide,Pyo3Docs,MiniTensorRepo}. The repository documents installation via \texttt{pip} and source builds with \texttt{maturin}, together with a small number of runtime requirements \citep{MiniTensorRepo,PyPI-MiniTensor}. Users can mix MiniTensor tensors with NumPy workflows because the API mirrors familiar shape and broadcasting semantics \citep{Harris2020}.

\subsection{Engine and Performance Techniques}

The Rust engine benefits from ahead of time compilation and LLVM vectorization. Inner loops in elementwise kernels and reductions are written to encourage auto vectorization. Where appropriate, the implementation can use portable SIMD abstractions that dispatch to ISA specific vector instructions on x86 or Arm \citep{RustSIMDGuide,RustSIMDcore}. Parallelism over independent chunks enables multi core scaling on large arrays. The engine also delays allocation of gradient buffers until a backward pass needs them. The repository documents these choices and marks GPU support as a roadmap item \citep{MiniTensorRepo}.

\section{Lightweight Footprint}

We quantify the size advantage using official wheels on PyPI.

\begin{table}[t]
\centering
\caption{Package sizes from PyPI at the time of writing. MiniTensor distributes a wheel of a few megabytes. PyTorch and TensorFlow wheels for common Linux or Windows targets are hundreds of megabytes. Sources: \citep{PyPI-MiniTensor,PyPI-Torch,PyPI-TensorFlow}.}
\label{tab:sizes}
\begin{tabular}{@{}lcc@{}}
\toprule
Package and platform & Wheel filename (example) & Size \\
\midrule
MiniTensor (Linux x86\_64, cp312) & \texttt{minitensor-0.1.1-...-.whl} & \textbf{2.6 MB} \\
PyTorch (Linux x86\_64, cp313t) & \texttt{torch-2.8.0-...-.whl} & \textbf{887.9 MB} \\
TensorFlow (Linux x86\_64, cp312) & \texttt{tensorflow-2.20.0-...-.whl} & \textbf{620.7 MB} \\
\bottomrule
\end{tabular}
\end{table}

Table~\ref{tab:sizes} shows that MiniTensor's binary distribution is extremely compact, which directly reduces download time and disk footprint. This difference arises from design choices. MiniTensor keeps the kernel surface minimal, leans on Rust's standard library, and avoids bundling large GPU backends in the default wheel. Researchers can still interoperate with NumPy and other Python tools while working inside a small, auditable codebase \citep{MiniTensorRepo,Harris2020}.

\section{Correctness and Testing}

MiniTensor includes unit tests for tensor arithmetic, broadcasting, autograd rules, and layer gradients \citep{MiniTensorRepo}. Reverse mode implementations can validate correctness by checking finite differences on random inputs, that is
\begin{equation}
\frac{\partial L}{\partial \theta_i} \;\approx\; \frac{L(\bm{\theta}+\epsilon \mathbf{e}_i)-L(\bm{\theta}-\epsilon \mathbf{e}_i)}{2\epsilon},
\end{equation}
with $\epsilon$ small. Although finite differences are slow, they provide a reference for edge cases and broadcasting semantics. The repository also demonstrates end to end examples that train small models and confirm consistent loss descent.

\section{Comparison with PyTorch and TensorFlow}

PyTorch and TensorFlow provide broad operator coverage, mature GPU and TPU backends, and distributed training stacks \citep{Paszke2019,Abadi2016}. MiniTensor does not attempt to replicate that scope. It focuses on a compact core that runs on CPUs with competitive constant factors for many elementwise operations and reductions, and it exposes an imperative autograd API that mirrors common research workflows \citep{MiniTensorRepo}. Users who require very large models, extensive operator sets, or multi device training should choose PyTorch or TensorFlow. Users who prioritize small binaries, ease of auditing, or teaching can adopt MiniTensor without the burden of heavy dependencies.

\section{Limitations and Roadmap}

MiniTensor currently supports dense tensors of 32 bit floats and a practical subset of neural network primitives. The public documentation marks GPU backends as a future enhancement and encourages contributions for advanced linear algebra and additional datatypes \citep{MiniTensorRepo}. The Python facing optimizer loops operate at the granularity of model parameters. If users encounter overhead in massive models, developers can migrate these loops into batched Rust kernels.

\section{Reproducibility and Availability}

Code, issues, tests, and examples live in the public repository \citep{MiniTensorRepo}. The PyPI page provides wheels for selected platforms and lists the minimal runtime requirements \citep{PyPI-MiniTensor}. Users can install with
\begin{quote}
\texttt{pip install minitensor}
\end{quote}
or build from source using \texttt{maturin}.

\section{Conclusion}

We described MiniTensor, a compact tensor library that uses a Rust engine and PyO3 bindings to deliver a clear Python API with reverse mode automatic differentiation. The library implements core layers and optimizers with mathematically standard derivatives and a small, auditable codebase. The published wheel size demonstrates a substantial footprint advantage over full scale frameworks while preserving the essentials for research and education on CPUs. Future work will broaden operator coverage and evaluate GPU backends.

\acks{We thank the open source communities of PyTorch, TensorFlow, Rust, and PyO3 for the foundational tools and documentation that inform this work. No fundings have been provided and no competing interests.}

\bibliography{minitensorbib}

\end{document}